\documentclass[sigconf,nonacm]{acmart}

\usepackage{graphicx}
\usepackage{comment}
\usepackage{amsmath,amssymb} 
\usepackage{color}
\usepackage{booktabs}
\usepackage{algorithm}
\usepackage{siunitx}
\usepackage[noend]{algpseudocode}
\usepackage{wrapfig,lipsum,booktabs}
\usepackage{enumerate}   
\usepackage{helvet} 
\usepackage{courier}  
\usepackage[utf8]{inputenc} 
\usepackage[T1]{fontenc}    
\usepackage{hyperref}

\usepackage{booktabs}       
\usepackage{amsfonts}       
\usepackage{nicefrac}       
\usepackage{microtype}      

\usepackage{epsfig}
\usepackage{amsmath}
\usepackage{amssymb}
\usepackage{adjustbox}

\usepackage{amsthm}  
\usepackage{wasysym}
\usepackage[ruled,vlined,algo2e]{algorithm2e}
\providecommand{\SetAlgoLined}{\SetLine}
\usepackage{color}
\usepackage{dsfont}	
\usepackage{wrapfig}
\usepackage{footnote}
\usepackage{epstopdf}
 \usepackage{enumitem}
\usepackage{caption}
\usepackage{subcaption}
\usepackage{comment}
\usepackage{multirow}

\def\ie{\emph{i.e., }}

\def\etal{\emph{et al. }}

\def\lidar{LiDAR}
\def\lodo{LodoNet}

\usepackage{pifont}

\DeclareMathOperator*{\argmin}{arg\,min}

\usepackage{mathtools}

\makeatletter
\newcommand*{\rom}[1]{\expandafter\@slowromancap\romannumeral #1@}
\makeatother
\makeatletter
\newcommand\footnoteref[1]{\protected@xdef\@thefnmark{\ref{#1}}\@footnotemark}
\makeatother
\newcommand{\bfsection}[1]{\vspace*{0.1cm}\noindent\textbf{#1.}}

\AtBeginDocument{%
  \providecommand\BibTeX{{%
    \normalfont B\kern-0.5em{\scshape i\kern-0.25em b}\kern-0.8em\TeX}}}

\begin{document}
\title{\lodo{}: A Deep Neural Network with 2D Keypoint Matching for 3D LiDAR Odometry Estimation}

\author{Ce Zheng}
\authornote{This work was done when the author visited VISLab \url{https://zhang-vislab.github.io}}
\email{czheng6@uncc.edu}
\affiliation{%
  \institution{The University of North Carolina at Charlotte}
}

\author{Yecheng Lyu}
\email{ylyu@wpi.edu}
\affiliation{%
  \institution{Worcester Polytechnic Institute}
}

\author{Ming Li}
\email{mli12@wpi.edu}
\affiliation{%
  \institution{Worcester Polytechnic Institute}
}

\author{Ziming Zhang}
\authornote{Corresponding author}
\email{zzhang15@wpi.edu}
\affiliation{%
  \institution{Worcester Polytechnic Institute}
}


\begin{abstract}
Deep learning based LiDAR odometry (LO) estimation attracts increasing research interests in the field of autonomous driving and robotics. Existing works feed consecutive LiDAR frames into neural networks as point clouds and match pairs in the learned feature space. In contrast, motivated by the success of image based feature extractors, we propose to transfer the LiDAR frames to image space and reformulate the problem as image feature extraction. With the help of scale-invariant feature transform (SIFT) for feature extraction, we are able to generate matched keypoint pairs (MKPs) that can be precisely returned to the 3D space. A convolutional neural network pipeline is designed for LiDAR odometry estimation by extracted MKPs. The proposed scheme, namely \lodo{}, is then evaluated in the KITTI odometry estimation benchmark, achieving on par with or even better results than the state-of-the-art.

\end{abstract}





\maketitle

\section{Introduction}

Odometry estimation is one of the key components in the automated driving systems and robotics. In recent years, autonomous driving has attracted significant research interests and several works have been proposed to estimate the position and orientation of the autonomous vehicles. Multiple sensors are deployed to collect relative data including monocular cameras, inertial measurement units (IMU), and light detection and ranging (LiDAR). Camera is first selected for odometry estimation because it is cost efficient and there exist well-studied image feature extractors to support. However, camera based methods such as VINS-Mono\cite{VINS-Mono}, DVSO\cite{DVSO}, and GANVO\cite{almalioglu2019ganvo} highly subject to the accuracy of camera calibration, and it is difficult to transfer from image feature pairs to coordinate correspondents between frames. Comparing with cameras, LiDARs take advantage of accurate distance acquisition and insensitive to the light condition. Therefore, LiDAR odometry (LO) attracts increasing research interests. Several works have been proposed to solve the vehicle odometry estimation using LiDAR data. They mainly follow two approaches. The first one performs traditional set-to-set registration that extracts features from the entire point cloud and register one to another. This approach includes LOAM\cite{LOAM} and ICP\cite{ICP_l}\cite{GICP}\cite{ICP_p2plane}. However, those approach only focus on the global features without capturing the local pattern, which limits its accuracy. The other approach use neural networks to extract local and global features and try to estimate the vehicle odometry through matching the feature vectors. However, Estimating odometry directly from large amount of 3D points is not only challenging but also computationally expensive. Motivated by the success of image feature extractors, we intuitively raise the question: 

\emph{Can we extract features from the LiDAR point clouds using image feature extractors so that the coordinate correspondences can be established through matching points in image feature space?}

Fortunately, Lyu \etal{} \cite{lyu2018real,lyu2018chipnet} and RangeNet++ \cite{milioto2019rangenet++} have introduced an algorithm to project the LiDAR data on to a spherical view so that a LiDAR point cloud with geometry features can be transferred to an image-like feature map with minor point losses. By employing this projection, we can efficiently generate image representations of LiDAR frames for feature extraction.

\begin{figure*}[t!]
\setlength{\belowcaptionskip}{-0.1cm} 
     \centering
     \begin{subfigure}[t]{0.3\textwidth}
         \centering
         \includegraphics[width=\textwidth]{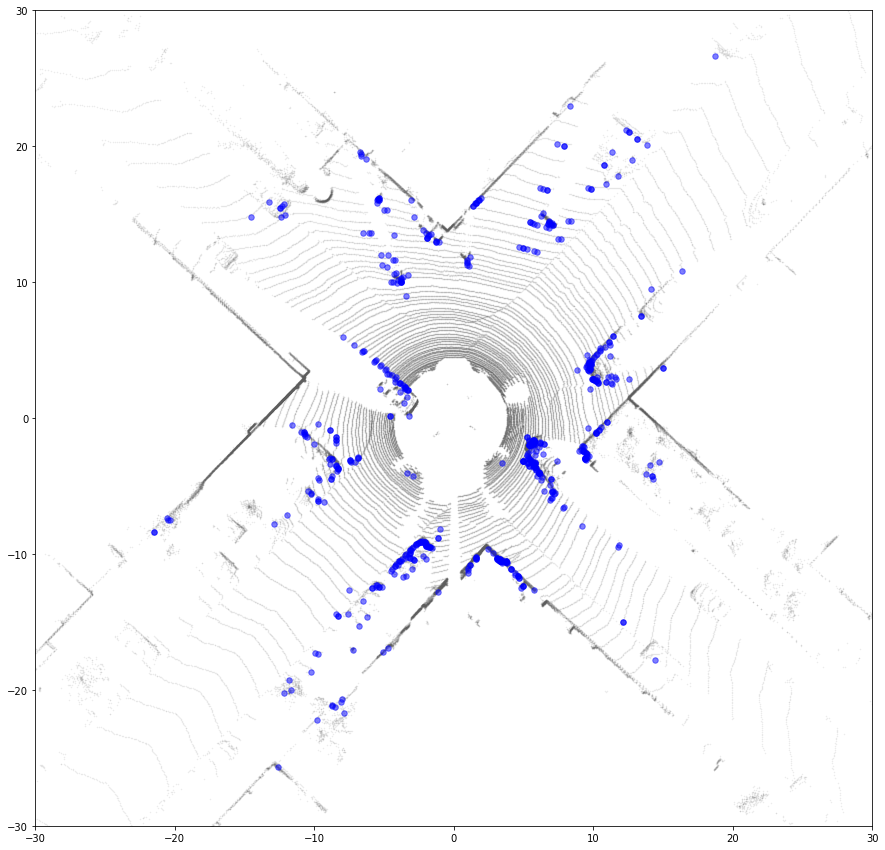}
         \captionsetup{font={small}} 
         \caption{}
         \label{fig:matching a}
     \end{subfigure}
     \begin{subfigure}[t]{0.3\textwidth}
         \centering
         \includegraphics[width=\textwidth]{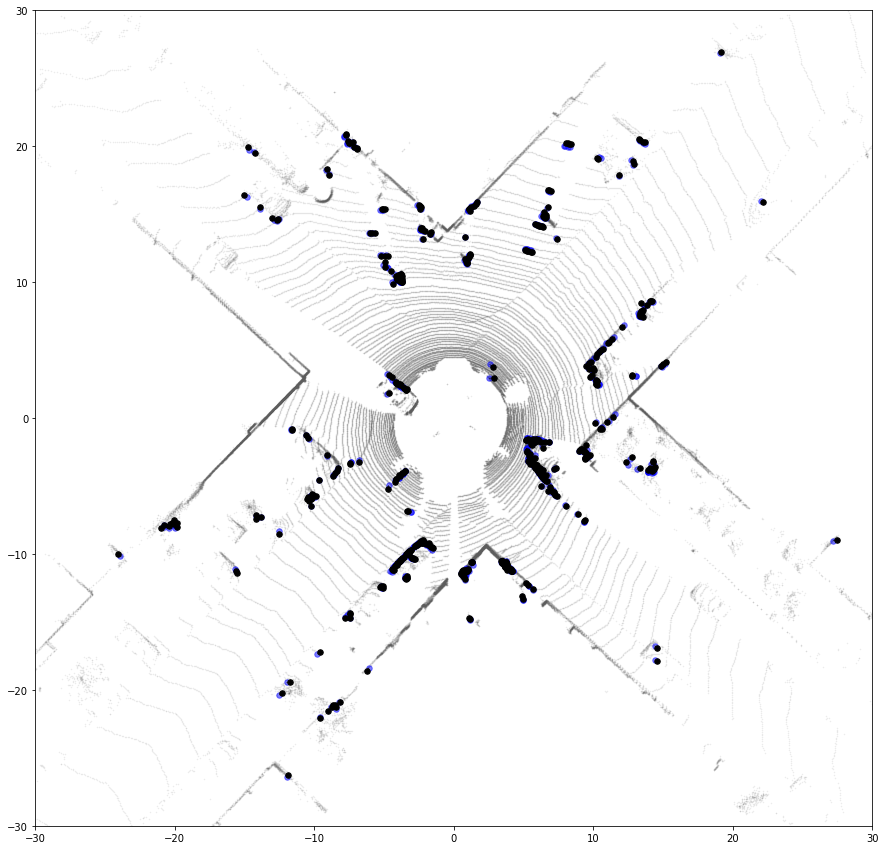}
         \captionsetup{font={small}} 
         \caption{}
         \label{fig:matching b}
     \end{subfigure}
     
        \caption{\footnotesize {Illustration of MKPs detected by our methods. (a): The point cloud of $i+1$'s LiDAR scan, blue points are the keypoints detected by SIFT on the coordinate frame of  $i+1$'s scan. (b): Black points are the MKPs of the blue points from the $i's$ scan and projected to the $i+1's$ coordinate frame. The black points are expected to overlap to their paired blue points since they are match points projected to the same coordinate frame. Lines are aligned for each matched pair (blue point and paired black point). Here these lines are invisible because of overlapping. }}
        \label{fig:matching3D}
\end{figure*}

\begin{figure*}[h]
  \centering
  \includegraphics[width=0.99\textwidth]{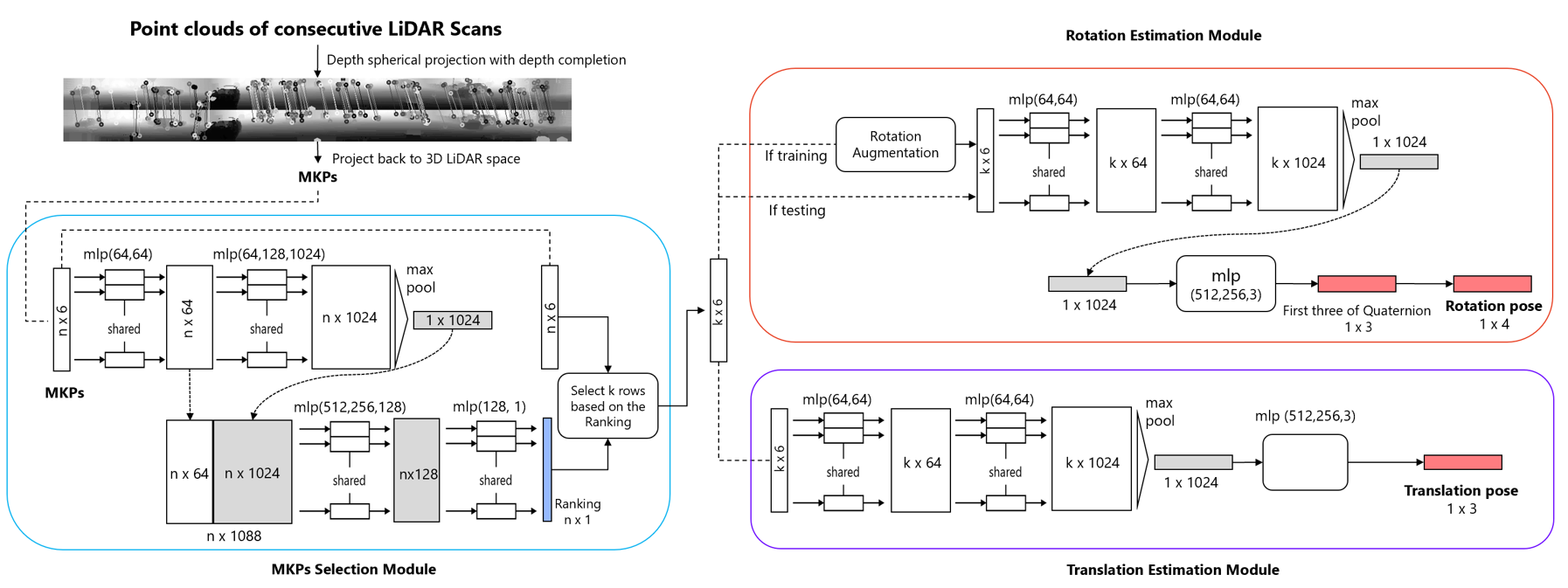}
  \vspace{-1em}
  \caption{\footnotesize{\lodo{} Architecture:Depth spherical images of the LiDAR point clouds can be obtained by the projection procedure. The matching keypoint pairs (MKPs) is extracted by the feature extraction algorithm:SIFT on the depth images than projected back to 3D LiDAR space. The MKPs Selection Module takes $n$ MKPs of consecutive scans as input. It can generate a Ranking Matrix $[n, 1]$ of the input $n$ MKPs, then we choose the best $k$ MKPs by their ranking as the input for rotation and translation estimation. The Rotation Estimation Module and Translation Estimation Module predict the rotation pose and translation pose respectively.Dotted line indicate two blocks are identical.}}
  \label{fig:network}
\end{figure*}

In this paper, we propose a deep neural network architecture to estimate the vehicle odometry from LiDAR frames. we first project the sparse LiDAR point clouds to spherical depth images with depth completion to tackle the sparse issue. We than apply the classic keypoints detection and matching algorithm to the 2D spherical depth images. The extracted matched keypoint pairs(MKPs) on 2D spherical image space can be projected back to the 3D LiDAR space by the inverse projection function. Figure \ref{fig:matching3D} illustrates that the MKPs extracted by our method are accurate and reliable on 3D LiDAR space. Inspired by the recent works on deep learning-based Visual Odometry (VO) estimation and the promising performance of PointNet\cite{qi2017pointnet} on point cloud segmentation and classification, we construct our deep neural network architecture to estimate LiDAR odometry using extracted MKPs as illustrated in Figure \ref{fig:network}.   

To summarize, our main contributions are:
\begin{itemize}
\item We propose a new approach for extracting matched keypoint pairs(MKPs) of consecutive LiDAR scans by projecting 3D point cloud onto 2D spherical depth images where MKPs can be extracted effectively and efficiently. After projecting back to 3D LiDAR space, the extracted MKPs can be used for LiDAR odometry estimation. 
\item By utilizing the PointNet structure, which provide strong performance over point cloud tasks, We adopt our convolutional neural network architecture to infer the rotation information and translation information from extracted MKPs of consecutive scans.
\item The evaluation of our experiments and ablation studies on KITTI Odometry estimation benchmark\cite{kitti} demonstrate the effectiveness of the proposed method. 
\end{itemize}

\section{Related work}
\bfsection{LiDAR feature extraction in 2D space}

LiDAR feature extraction is one of the essential tasks of LiDAR-based applications. Since the raw data from the LiDAR sensors are not well structuralized and ordered, an intuitive way is to project the LiDAR points onto a 2D space. LoDNN \cite{caltagirone2017Lodnn} is a pioneering work that projects all LiDAR points onto the ground plane and samples them to an image-like array. In this work, however, the LiDAR points are not uniformly distributed on the ground plane but heavily gathered together near the LiDAR scanner, which results in massive dropped points in the near-range and redundant space in the far-range. Lyu \etal{} \cite{lyu2018real,lyu2018chipnet} and RangeNet++ \cite{milioto2019rangenet++} improve the projection scheme by replacing the target plane with a sphere surface, in which LiDAR points are nearly uniformly distributed. SqueezeSeg V1 \cite{wu2018squeezeseg}, V2 \cite{wu2019squeezesegv2}, and LO-Net \cite{LO-Net} also employ this projection scheme and result in a good performance in LiDAR point semantic segmentation. In our work, we follow the projection scheme of RangeNet++ to generate a 2D LiDAR feature map for each LiDAR frame.

\bfsection{Image feature extraction and keypoint detection}

Image feature extraction and keypoint detection have been studied for decades. Scale-invariant feature transform (SIFT) \cite{lowe1999sift} is one of the popular feature extraction algorithms in the image processing domain. By transforming an image into a large group of feature vectors that are invariant to image translation, scaling, and rotation, SIFT generates robust feature vectors on parts captured in different views. Other feature extraction methods include SURF \cite{bay2006surf} and ORB \cite{rublee2011orb}, however, they cannot generate robust feature vectors on LiDAR frames. In our work, we employ the SIFT as our LiDAR feature extractor.

\bfsection{Registration and feature-based vehicle odometry estimation}

Iterative Closet Point (ICP)\cite{ICP} method and its variants\cite{ICP_variants}\cite{ICP_variants1}\cite{ICP_variants2} have been used in the field of LiDAR based pose estimation widely. In the ICP algorithm, a transformation between point clouds of adjacent scans is optimized iteratively by minimizing distance until a specific termination condition is met. Despite its popularity, ICP is sensitive to the initial poses and computationally expensive. Multiple variants of ICP algorithms such as point-to-line ICP\cite{ICP_l}, point-to-plane ICP\cite{ICP_p2plane}, and plane-to-pane ICP\cite{ICP_variants} were developed. GICP\cite{GICP} was proposed by combining point-to-point ICP and point-to-plane ICP into a single probabilistic framework. A probabilistic model is attached to the minimization step which can reduce the time complexity and increase the robustness to incorrect correspondences. Collar Line Segments (CLS)\cite{CLS} transforms the LiDAR scans into line could then generates line segments within each bin. This pre-processing method produces better results than GICP. However, due to the high computational cost in line segments, CLS cannot achieve real-time odometry estimation. 

The state-of-art LiDAR Odometry estimation method: LiDAR Odometry And Mapping(LOAM)\cite{LOAM} was proposed by Zhang and Singh, which achieves both low-drift in motion estimation and low-computational complexity. The key idea is to divide the complex problem of Simultaneous Localization and Mapping into two parallel algorithms. One algorithm performs odometry at a high frequency but at low fidelity to estimate the velocity of the laser scanner. A second algorithm runs at an order of magnitude lower frequency for fine matching and registration of the point cloud.

\bfsection{Deep Learning-based vehicle odometry estimation}

In recent years, several works have been done exploring the use of neural networks in vehicle odometry estimation. DeepVO\cite{DeepVO}, DVSO\cite{DVSO}, Depth-VO-Feat\cite{Depth-VO-Feat}, and GANVO\cite{almalioglu2019ganvo} have achieved promising results on Visual odometry(VO) estimation. In VO tasks, Camera data are used. However, applying deep learning method to solve 3D LiDAR odometry problem still remains challenges. DeepICP\cite{Deep_icp} detects the keypoints by a point weighting layer, and then generates a search region for each keypoint. A matched point can be generated by the corresponding point generation layer. The odometry is finally estimated by solving the SVD from the matched keypoint pairs. DeepPCO\cite{DeepPCO} generates panoramic-view of depth image projection to feed to it neural networks. L3-Net \cite{lu2019l3} proposes a learning-based LiDAR localization system by comparing the network-based feature vector between the current LiDAR frame and pre-build point cloud map followed by a recurrent neural network based smoothness module. LO-Net\cite{LO-Net} is another learning-based odometry estimator. Different from L3-Net that only extracts point-wise features, LO-Net  projects the points onto a sphere surface and builds an image-like feature map for each LiDAR frame. For better training the odometry estimator, LO-Net introduces an attention branch to predict if the geometric consistency of an area in the feature map can be modeled or not.

\section{Method}
\subsection{Problem Setup}

We are given a collection of training samples $\{(\mathcal{X}_t, \mathcal{X}_{t+1}, \mathbf{y}_{t})\}_{t\in\mathcal{T}}$ where $\mathcal{X}_t\subseteq\mathbb{R}^3$ denotes the set of keypoints from the $t$-th \lidar{} scans, $\mathbf{y}_{t}\in\mathcal{Y}\subseteq\mathbb{R}^6$ denotes the ground-truth odometry between the $t$-th and $(t+1)$-th \lidar{} scans, and $\mathcal{T}$ denotes the scan index set. Our goal is to learn an odometry prediction function $f:\mathcal{X}\times\mathcal{X}\rightarrow\mathcal{Y}\in\mathcal{F}$ by minimizing certain loss function $\ell:\mathcal{Y}\times\mathcal{Y}\rightarrow\mathbb{R}$, \ie
\begin{align}\label{eqn:general}
    \min_{f\in\mathcal{F}}\sum_{t\in\mathcal{T}}\ell(f(\mathcal{X}_t, \mathcal{X}_{t+1}), \mathbf{y}_{t}).
\end{align}
Note that Eqn. \ref{eqn:general} holds in general for all the \lidar{} odometry estimation algorithms. To simplify our explanation later, here we assume that the feasible space $\mathcal{F}$ is proper, closed, and convex (PCC) that covers all the constraints on $f$ such as regularization. At test time, given two sets of keypoints $\mathcal{X}_{t'}, \mathcal{X}_{t'+1}$, we can predict their odometry as $f(\mathcal{X}_{t'}, \mathcal{X}_{t'+1})$.

\subsection{Formulation}

As we know, odometry has 6 degrees of freedom (DOF). This leads to the fact that odometry can be estimated given at least 3 matched keypoint pairs (MKPs). Therefore, instead of learning the general function $f$ in Eqn. \ref{eqn:general} directly, in the literature it will be more plausible to decompose it as two functions, \ie $f=h\circ g$ where $\circ$ denotes the function composition. The keypoint matching function, $g:\mathcal{X}_t\times\mathcal{X}_{t-1}\rightarrow\mathcal{P}_{t}\times\mathcal{P}_{t-1}\subseteq\mathbb{R}^3\times\mathbb{R}^3$, generates MKPs from the input keypoint sets, and the odometry regression function, $h:\mathcal{P}_t\times\mathcal{P}_{t-1}\rightarrow\mathcal{Y}$, predicts the odometry based on the MKPs. Odometry estimation algorithms are all about how to design or learn such functions $g, h$ to determine function $f$.

\bfsection{ICP-based learning approaches}
Iterative Closest Point (ICP) \cite{besl1992method} matches two sets of points iteratively as well as estimating the pose transformation by minimizing distances between the corresponding matched points until it converges. Although ICP is well-known for point registration, the high computation and sensitivity to initial poses significantly limit its applications. The key idea behind ICP-based learning approaches for \lidar{} odometry estimation is to use ICP to locate MKPs (given the current odometry regression function $h$) that are used to update $h$ further. In general, such approaches can be formulated as follows:
\begin{align}\label{eqn:ICP}
\small
    \min_{h\in\mathcal{H}} \sum_{t\in\mathcal{T}}\ell(h(\mathcal{P}_{t}, \mathcal{P}_{t+1}), \mathbf{y}_{t}), 
    s.t. \; \mathcal{P}_{t}, \mathcal{P}_{t+1} = \Tilde{g}(\mathcal{X}_{t}, \mathcal{X}_{t+1}, h(\mathcal{P}_{t}, \mathcal{P}_{t+1})) 
\end{align}
where $\tilde{g}:\mathcal{X}\times\mathcal{X}\times\mathcal{Y}\rightarrow\mathcal{P}\times\mathcal{P}$ denotes a variant of the keypoint matching function, and the feasible space $\mathcal{H}$ is PCC. 

The constraint here models the MKPs as the stationary solution given current $h$, which can be viewed as a generalization of ICP. Then such solutions are used to update $h$ in the objective by minimizing the loss. At training time this procedure is repeated until it converges. At test time, given $\tilde{g}$ and learned $h$ we use the constraint to locate the MKPs and then output the odometry estimation as $h(\mathcal{P}_{t'}, \mathcal{P}_{t'+1})$. 

\begin{align}\label{eqn:ICP_PQ}
\small
    \mathbf{Q}_i=
    \begin{bmatrix}
        x_{i,1}^2+x_{i,2}^2 & -x_{i,2}x_{i,3} & -x_{i,1}x_{i,3} & x_{i,2} & -x_{i,1} & 0 \\
        -x_{i,2}x_{i,3} & x_{i,1}^2+x_{i,3}^2 & -x_{i,1}x_{i,2} & -x_{i,3} & 0 & x_{i,1} \\
        -x_{i,1}x_{i,3} & -x_{i,1}x_{i,2} & x_{i,2}^2+x_{i,3}^2 & 0 & x_{i,3} & -x_{i,2} \\ 
        x_{i,2} & -x_{i,3} & 0 & 1 & 0 & 0 \\
        -x_{i,1} & 0 & x_{i,3} & 0 & 1 & 0 \\
        0 & x_{i,1} & -x_{i,2} & 0 & 0 & 1
    \end{bmatrix}
\end{align}

\begin{align}
\small
    \mathbf{q}_i = 
    \begin{bmatrix}
        y_{i,1}x_{i,2} - y_{i,2}x_{i,1} \\
        -y_{i,1}x_{i,3} + y_{i,3}x_{i,1} \\
        y_{i,2}x_{i,3} - y_{i,3}x_{i,2} \\
        x_{i,1} - y_{i,1} \\
        x_{i,2} - y_{i,2} \\
        x_{i,3} - y_{i,3}
    \end{bmatrix}
\end{align}

\begin{align}
\small
    \begin{bmatrix}
        \alpha,
        \beta,
        \gamma,
        b_1,
        b_2,
        b_3
    \end{bmatrix}^\top=-\left[\sum_i\mathbf{Q}_i\right]^{-1}\sum_i\mathbf{q}_i
\end{align}

\bfsection{Our \lodo{}}
As we see in Eqn. \ref{eqn:ICP}, the odometry regression function $h$ and the keypoint matching function $g$ in ICP-based learning approaches are essentially {\em coupled}. This potentially can lead to two serious problems, at least, in training, \ie high computation and non-convergence of the training loss.

To address such problems, our methodology in \lodo{} is to decouple functions $g, h$ to avoid the loop as well as significantly improve the convergence. To this end, we propose the following optimization problem:
\begin{align}\label{eqn:lodonet}
\small
    \min_{z\in\mathcal{Z},  \Tilde{h}\in\tilde{\mathcal{H}}}  \sum_{t\in\mathcal{T}}\ell & (\tilde{h}(\mathcal{P}_{t},  \mathcal{P}_{t+1}, z(\mathcal{P}_{t}, \mathcal{P}_{t+1})), \mathbf{y}_{t}), \; \notag \\
    s.t. \;& \mathcal{P}_{t}, \mathcal{P}_{t+1} = g(\mathcal{X}_{t}, \mathcal{X}_{t+1}), 
\end{align}
where $z:\mathcal{P}\times\mathcal{P}\rightarrow\Pi$ denotes an {\em attentional} function that returns probability vectors over the MKPs in the simplex space $\Pi$, $\tilde{h}:\mathcal{P}\times\mathcal{P}\times\Pi\rightarrow\mathcal{Y}$ denotes a variant of the odometry regression function, and both spaces $\mathcal{Z}, \tilde{H}$ are PCC.

Different from ICP-based learning approaches, here we use a predefined keypoint matching function $g$ to extract MKPs from \lidar{} data (\ie constraint), and feed these MKPs as input to the learning algorithm directly to minimize the objective. In this way, there is no loop between $g$ and $\tilde{h}$ (\ie decoupling). The quality of the MKPs, however, cannot be guaranteed to be good for odometry estimation. Therefore, we deliberately introduce the attentional mechanism to assign weights for estimation. In fact, Eqn. \ref{eqn:lodonet} is an unconstrained minimization problem with convergence guarantee using alternating optimization (\ie learning $z$ while fixing $\tilde{h}$, and then learning $\tilde{h}$ while fixing $z$), in general, as $(\mathcal{P}_t, \mathcal{P}_{t+1}), \forall t\in\mathcal{T}$ now are the inputs. At test time we predict the odometry as $\tilde{h}(\mathcal{P}_{t}, \mathcal{P}_{t+1}, z(\mathcal{P}_{t}, \mathcal{P}_{t+1}))$.

\subsection{Data prepossessing}
\label{section:Data prepossessing}


\begin{figure*}[h]
     \centering
     
     \begin{subfigure}[b]{0.9\textwidth}
         \centering
         \includegraphics[width=0.4\textwidth]{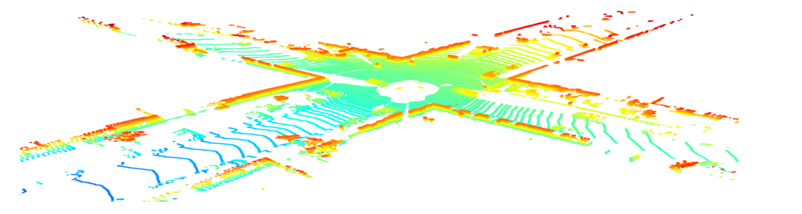}
         \captionsetup{font={scriptsize}} 
         \caption{Point Cloud of one LiDAR scan}
         \label{fig:spherical a}
     \end{subfigure}
    
     \begin{subfigure}[b]{0.95\textwidth}
         \centering
         \includegraphics[width=\textwidth]{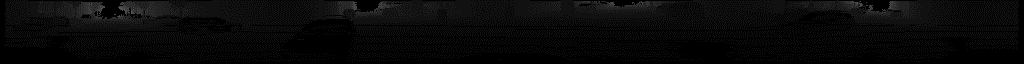}
         \captionsetup{font={scriptsize}} 
         \caption{Original spherical image of one scan of LiDAR Point}
         \label{fig:spherical b}
     \end{subfigure}
     
     \begin{subfigure}[b]{0.95\textwidth}
         \centering
         \includegraphics[width=\textwidth]{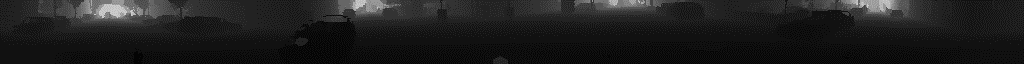}
         \captionsetup{font={scriptsize}} 
         \caption{Depth spherical image of the original spherical image}
         \label{fig:spherical c}
     \end{subfigure}

     \begin{subfigure}[b]{0.95\textwidth}
         \centering
         \includegraphics[width=\textwidth]{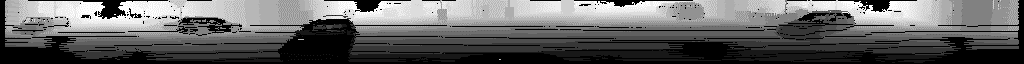}
         \captionsetup{font={scriptsize}} 
         \caption{Histogram equalization of the original spherical image}
         \label{fig:spherical d}
     \end{subfigure}
     
     \begin{subfigure}[b]{0.95\textwidth}
         \centering
         \includegraphics[width=\textwidth]{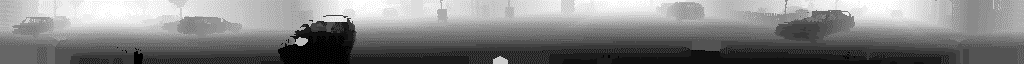}
         \captionsetup{font={scriptsize}} 
         \caption{Histogram equalization of the depth completion spherical image}
         \label{fig:spherical e}
     \end{subfigure}
     
        \caption{The spherical image of one LiDAR scan}
        \label{fig:spherical}
\end{figure*}

\bfsection{Depth spherical image generation from LiDAR point clouds}

The 3D LiDAR point clouds are usually stored by a set of Cartesian coordinates $(X, Y, Z)$. Due to the relative low resolution of LiDAR scanners, the 3D LiDAR point clouds are quite sparse. A scan of the LiDAR point cloud is shown in Figure \ref{fig:spherical a}. However, matching keypoints pairs from two consecutive scans of LiDAR point clouds would be inaccurate and time/memory consuming. Therefore, we project the sparse LiDAR point clouds to spherical projection images that are nearly dense. The projection function is: 

\begin{align}
\small
    \alpha = \arcsin(\frac{z}{\sqrt{x^2 + y^2 +z^2}}), \; 
    \Bar{\alpha} = | \frac{\alpha}{	\Delta\alpha} |
\end{align}

\begin{align}
\small
    \beta = \arcsin(\frac{y}{\sqrt{x^2 + y^2}}), \;
    \Bar{\beta} = | \frac{\beta}{	\Delta\beta} |
\end{align}
Where $\alpha$ and $\beta$ are the indexes of points' position in the matrix. $\Delta\alpha$ and $\Delta\beta$ are the angular resolutions in the horizontal and vertical directions, respectively. The element at $(\alpha, \beta)$ of the spherical image is set to be the range value $r = \sqrt{x^2 + y^2 +z^2} $ of the LiDAR point $(x, y, z)$ \cite{LO-Net}. A matrix of size $H \times W \times C$ can be obtained by applying this projection. $H$ is the number of vertical channels from LiDAR scanners, $W$ is the width of the spherical image, and $C$ is the channel of input point cloud matrix. Figure \ref{fig:spherical b} shows the spherical image of the LiDAR scan in Fig \ref{fig:spherical a}.

\bfsection{Depth completion and histogram equalization}

In depth completion we aim to fill the void pixels in our projected spherical image, so that every pixel can be utilized in the following algorithms. Traditional image inpainting algorithms try to interpolate the target pixels using surrounding pixel color values, which is computational expensive for real-time processing. In this paper, based on the assumption that depth value does not change much in local regions, we speed up the depth completion by filling the void pixels with the depth value of its nearest valid pixels. The algorithm is described in algorithm \ref{alg:Depth Completion}. 
\begin{algorithm}[t]\footnotesize
	\SetAlgoLined
	\begin{flushleft}

     \textbf{INPUT:} valid pixel set $\mathcal{P}_1 = \{\mathbf{p}_1\}$, void pixel set $\mathcal{P}_0 = \{\mathbf{p}_0\}$\\
	 \textbf{OUTPUT:} filled pixel set $\mathcal{P}_0 = \{\mathbf{p}_0\}$\\
	\end{flushleft}
    \begin{algorithmic}[1]
    \State \textbf{For Each} $\mathbf{p}_0$ in $ \mathcal{P}_0 $:
    \State \quad  $\mathbf{p}' \in \argmin_{\mathbf{p}_1\in \mathcal{P}_1}\left\{||\mathbf{p}_0 - \mathbf{p}_1||\right\}$
,   \State \quad  $\mathbf{p}_0 \leftarrow \mathbf{p}^*$
    \State \textbf{End}
    \State Return $\mathcal{P}_0$
    
    \end{algorithmic}
	\caption{\footnotesize Depth Completion}\label{alg:Depth Completion}
\end{algorithm}

As shown in the Figure ~\ref{fig:spherical b}, the value in each pixel represents the distance of the original LiDAR detected point to the LiDAR that causes the most pixels in the spherical image to have a relatively small value. To improve the spherical image’s visual quality, we apply the histogram equalization technique in order to enhance the contrast as shown in Figure ~\ref{fig:spherical c}. Histogram equalization is the most popular contrast enhancement technique due to its simplicity and effectiveness \cite{2008digital}. This widely used technique is achieved by flattening the dynamic range of an image’s histogram of grey value based on the probability density function \cite{contrastenhancement}. After we generate the depth completion spherical image in Figure ~\ref{fig:spherical d} , we apply the Histogram equalization. Therefore, the brightness of the depth completion spherical image is improved significantly as shown in the Figure ~\ref{fig:spherical e}.

\bfsection{Keypoints detection and matching}

After the depth completion and histogram equalization step, we want to detect a group of MKPs from consecutive frames of spherical images $f_i$ and $f_{i+1}$. For example, one keypoint is shown in $f_i$ at the location $(x_1, y_1)$  and its matching point is located at $({x_1}^\prime, {y_1}^\prime)$ in $f_{i+1}$. Since this MKP represents the same object in consecutive frames, the potential odometry information between these consecutive frames is related to their location in spherical images. Thus, we apply the SIFT algorithm to detect keypoints in consecutive frames of spherical images. This local feature extraction method can extract comprehensive keypoints which are invariant to the object translation and rotation. Each detected keypoint will be represented as a 128$-$element feature vector, called by descriptors \cite{whensift}. The feature vectors will be used for keypoint matching.

\begin{figure*}[h]
     \centering
     \begin{subfigure}[b]{0.9\textwidth}
         \centering
         \includegraphics[width=\textwidth]{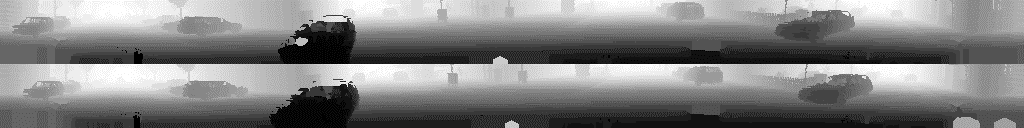}
         \captionsetup{font={small}} 
         \caption{2 consecutive frames of LiDAR scans}
     \end{subfigure}

     \begin{subfigure}[b]{0.9\textwidth}
         \centering
         \includegraphics[width=\textwidth]{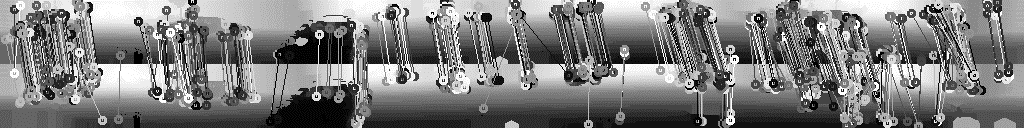}
         \captionsetup{font={small}} 
         \caption{Top 400 MKPs detected by SIFT}
     \end{subfigure}
     
     \begin{subfigure}[b]{0.9\textwidth}
         \centering
         \includegraphics[width=\textwidth]{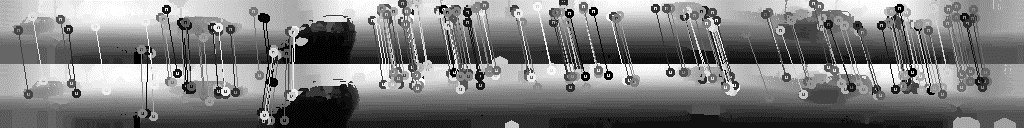}
         \captionsetup{font={small}} 
         \caption{100 MKPs selected by MKPs Selection Module}
     \end{subfigure}
     \vspace{-0.3cm}
        \caption{MKPs extracted from consecutive LiDAR scans}
        \label{fig:MKPs}
\end{figure*}

Given a keypoint from depth completion image  $f_i$, we want to find the best matching point in $f_{i+1}$ to form the MKP. The similarity will be measured by the Euclidean distance between keypoints in the spherical image \cite{siftvo}. In Figure ~\ref{fig:MKPs}a, two consecutive scans are concatenated vertically. The Figure ~\ref{fig:MKPs}b shows the top 400 MKPs detected from these consecutive scans. However, some MKPs are mismatched or not suitable for odometry estimation. For example, we do not want to use any point from dynamic objects. Figure ~\ref{fig:MKPs}c illustrates the 100 MKPs selected by MKP Selection module among the 1000 MKPs detected in this section. We will discuss this issue in section \ref{section:train lodo}.

\bfsection{Projecting back to 3D point cloud}

It is simple to project MKPs in the spherical image back to 3D point clouds. Noticed that the MKPs were detected from depth completion spherical image, some MKPs may not have corresponding points in the original 3D space. In other words, some “fake” MKPs are generated by depth completion, not from original LiDAR points. These “fake” MKPs are removed and only real MKPs return the points coordinate in 3D space. The MKPs which related to the odometry rotation and translation are extracted from entire LiDAR points of consecutive frames. Then we aligned the points in frame $f_i$ with their matching points in frame $f_{i+1}$ to construct a matrix with the size of $[m , 6]$ where $m$ is the number of MKPs we find, and 6 features represent the $x$, $y$, $z$ coordinates of matching points in consecutive frames. 

\subsection{LiDAR odometry regression}
\label{section:train lodo}
\bfsection{MKPs Selection module}

In Section \ref{section:Data prepossessing}, we extract MKPs of consecutive frames which contain the odometry rotation and translation information. However, some of MKPs are inappropriate for odometry estimation. We want to exclude those MKPs from dynamic objects such as moving cars. Even though matching are good, these MKPs may inhibit the odometry estimation due to the inconsistent odometry information of dynamic objects with LiDAR device. Hence a selection module is needed to determine whether a matching point should be used for odometry estimation. 

we deploy a MKPs Selection module to solve this segmentation problem in order to improve the effectiveness and robustness of the network. Here we use the PointNet segmentation structure to achieve the goal. The PointNet consists of symmetry function for unordered input, a local and global information aggregation, and a joint alignment network. In our MKPs Selection module, we modify several parts to fit our scenario. T-Net structure has been removed since we don't want to preserve the rotation invariant. The input are MKPs of consecutive frames, not original unordered points. For the output $M_S(P_i) \in [0,1]$, ground truth odometry which containing rotation $R$ and translation $T$ information are used to calculate the distance of the point in frame i with its matching point’s projection in frame i by the following equation: 
\begin{align}
\small
    d =\| \begin{bmatrix}
            R & T \\
            0 & 1 
            \end{bmatrix}  \mathcal{X}_i - \mathcal{X}_{i+1} \|_2 
\end{align}
The output is set as 1 if the distance is small while large distance indicates the output is 0. MKPs with label 1 will be use in the odometry estimation.

\bfsection{Rotation Estimation module}

\begin{algorithm}[t]\footnotesize

	\SetAlgoLined
	\begin{flushleft}
    \textbf{INPUT:} Point cloud pair $\mathcal{P}_1,\mathcal{P}_2$, odometry Matrix Ground Truth $Tr$, augmentation rotation upper-bound $beta_{max}$ \\
	\textbf{OUTPUT:} Augmented point cloud pair $\mathcal{P}'_1,\mathcal{P}'_2$, augmented odometry Matrix Ground Truth $Tr'$ \\ 
	\end{flushleft}
    \begin{algorithmic}
    \State$\beta \leftarrow Random(-\beta_{max},\beta_{max})$
    \State$Tr1 \leftarrow Yaw(\beta)$
    \State$Tr' \leftarrow [Tr]^{-1}\mathcal{P}_2$
    \State$\mathcal{P}'_1 \leftarrow \mathcal{P}_1$
    \State$\mathcal{P}'_2 \leftarrow Tr1 \mathcal{P}_1$
    \State Return $\mathcal{P}'_1,\mathcal{P}'_2,Tr'$
    \end{algorithmic}
	\caption{\footnotesize Data Augmentation on Odometry Rotation}\label{alg:augmentation}

\end{algorithm}
After selecting a fixed number of MKPs, we aim to infer the 3-DoF relative odometry rotation information by constructing a regression model. By observing popular autonomous driving datasets with LiDAR odometry task, recording cars usually go straight line while turning frames are not enough. Hence, we apply a data augmentation procedure for rotation estimation. Because each matching pair associate with the 3-DoF odometry rotation information R, the Eqn. \ref{eqn:ICP_PQ} represents the relation between matching pairs with a rotation matrix. We can generate the augmented MKPs with corresponding rotation matrix by given the existing MKPs with their rotation matrix based on the Algorithm \ref{alg:augmentation}.

The structure of rotation estimation module is based on the PointNet classification model while we change it to the regression model. We remove the T-net structure and add our rotation augmentation structure. The input of the Rotation Estimation module is selected MKPs from the MKPs Selection module concatenate with rotation augmentation of selected MKPs. The output is the Unit Quaternion format of the odometry rotation information. The Unit Quaternion of two consecutive frames can be represented by a vector $[a, bi, cj, dk]$, where a, b, c, and d are real numbers, and i, j, and k are the fundamental Quaternion units \cite{Quaternion}. Based on the norm of the Unit Quaternion vector is equal to 1, here we only predict the first three dimension of Unit Quaternion and the last dimension can be calculated afterwards. 
The rotation loss function of consecutive frames is defined as: 
\begin{align}
\small
    \mathcal{L}_R =\|  Q - \hat{Q} \|_l 
\end{align}
Where $Q$ is the ground truth first three dimension of Unit Quaternion, $\hat{Q}$ is the predicted first three dimension of Unit Quaternion by the network, and $l$ refers to the Euclidean distance norm. Here we choose the $l_2$ norm in this module. 

\bfsection{Translation Estimation module}

The network structure of Translation estimation module is quite similar to the Rotation Estimation module. In contrast, we directly apply the selected MKPs as input to feed the Translation Estimation module since the translation augmentation is not required. The output is the 3-DoF odometry translation information and the loss function is defined as: 
\begin{align}
\small
    \mathcal{L}_T =\|  T - \hat{T} \|_l 
\end{align}
Where $T$ is the ground truth translation matrix, $\hat{T}$ is the predicted translation array by the network, and $l$ refers to the Euclidean distance norm. Here we choose the $l_2$ norm in this module. 

\section{Experiments}
\subsection{Implementation detail}
\label{section:Implementation detail}
In our experiment, we use the point cloud data which is collected by the Velodyne HDL-64 3D LiDAR sensor to estimate the odometry. When converting the LiDAR point clouds to spherical images, we set the height of the image to 64 and the width to 1024. For each two consecutive scans $f_i$ and $f_{i+1}$, we detect MKPs on spherical images then convert back to 3D space. In a sequence of point clouds with $n+1$ scans, we can collect 1000 MKPs for $n$ consecutive frames to form an input matrix $[n, 1000, 6]$. \lodo{} predicts the odometry between n consecutive frames and give an output as $[n, 7]$, then we can calculate the odometry matrix $[n, 12]$ indicating each scan’s odometry by the relation of rotation and translation matrix. The whole framework is implemented with the Tensorflow, we choose the Adam optimizer\cite{adam} for optimization. The batch size and learning rate are set to 128 and 0.0001. MKPs Selection module select 100 MKPs among 1000 MKPs for each consecutive scans. One GeForce RTX 2080 Ti GPU is used for training. 

\subsection{Dataset}

\quad \textbf{KITTI}. The KITTI odometry dataset\cite{kitti} is a popular and widely used dataset in the autonomous driving tasks. It provides 22 independent sequences with stereo gray-scale camera images, color camera images, and point clouds captured by a Velodyne HDL-32 LiDAR sensor among urban, highway and countryside scenes. For our work, we use Velodyne LiDAR data only to estimate odometry. For sequence 0 to sequence 10, KITTI dataset provides the ground truth odometry, while for sequence 11 to 21 the ground truth is preserved for online benchmark testing. Hence, we use sequence 0, 1, 2, 3, 4, 5, 6, 9, and 10 to train our model while leaving sequence 7 and sequence 8 for validation.

As we state in section \ref{section:train lodo} that odometry has 6 degrees of freedom which can be represented by rotation and translation components. In the ground-truth odometry pose files provided by KITTI dataset, a 12 by 1 vector is assigned for the odometry pose of $i_{th}$ frame to the 0 frame while 9 of them indicating the rotation and 3 of them present the translation. This 9-dimension rotation vector usually reshapes to a 3 by 3 rotation matrix. In our experiments, we convert this 3 by 3 rotation matrix to a 1 by 4 Unit Quaternion representation due to less dimension for network training. Noted here we only predict first three dimension of Unit Quaternion and the last dimension can be calculated directly since the norm of the Unit Quaternion is equal to 1. Thus, given the input of the MKPs of two consecutive frames, the output of our network is a 1 by 7 vector which is concatenated by a 1 by 4 rotation vector with a 1 by 3 translation vector.

\subsection{Evaluation}
We compared our method with the ground truth trajectory and several LiDAR odometry estimation methods: ICP-point2point(ICP-po2po), ICP-point2plane (ICP-po2pl), GICP\cite{GICP}, CLS\cite{CLS}, LOAM\cite{LOAM}, Velas et al.\cite{Velas}, LO-Net\cite{LO-Net}.

\begin{table*}
\centering
  \caption{Odometry results on KITTI dataset}
  \resizebox{0.9\textwidth}{!}{
\begin{tabular}{|c|cccccccccccccccccc|cc|}
\hline
                      & \multicolumn{2}{c}{ICP-po2po} & \multicolumn{2}{c}{ICP-po2pl} & \multicolumn{2}{c}{GICP\cite{GICP}} & \multicolumn{2}{c}{CLS\cite{CLS}} & \multicolumn{2}{c}{LOAM\cite{LOAM}}      & \multicolumn{2}{c}{Velas et al\cite{Velas}} & \multicolumn{2}{c}{DeepPCO\cite{DeepPCO}} & \multicolumn{2}{c}{LO-Net\cite{LO-Net}} & \multicolumn{2}{c|}{ours}                                 \\ \cline{2-19} 
\multirow{-2}{*}{seq} & t\_rel        & r\_rel        & t\_rel        & r\_rel        & t\_rel  & r\_rel         & t\_rel     & r\_rel     & t\_rel        & r\_rel        & t\_rel         & r\_rel         & t\_rel       & r\_rel       & t\_rel          & r\_rel   & t\_rel                      & r\_rel                      \\ \hline
00                    & 6.88          & 2.99          & 3.80          & 1.73          & {\color[HTML]{3531FF} 1.29}    & {\color[HTML]{3531FF} 0.64}           & 2.11       & 0.95       & \textbf{0.78} & \textbf{0.53} & 3.02           & /              & /            & /            & 1.47            & 0.72     & 1.43 & 0.69 \\
01                    & 11.21         & 2.58          & 13.53         & 2.58          & 4.39    & 0.91           & 4.22       & 1.05       & 1.43          & 0.55          & 4.44           & /              & /            & /            & {\color[HTML]{3531FF} 1.36}            & {\color[HTML]{3531FF} 0.47}    & \textbf{0.96}               & \textbf{0.28}               \\
02                    & 8.21          & 3.39          & 9.00          & 2.74          & 2.53    & 0.77           & 2.29       & 0.86       & \textbf{0.92} & \textbf{0.55} & 3.42           & /              & /            & /            & 1.52            & 0.71     & {\color[HTML]{3531FF} 1.46} & {\color[HTML]{3531FF} 0.57} \\
03                    & 11.07         & 5.05          & 2.72          & 1.63          & 1.68    & 1.08           & 1.63       & 1.09       & \textbf{0.86} & \textbf{0.65} & 4.94           & /              & /            & /            & {\color[HTML]{3531FF} 1.03}            & {\color[HTML]{3531FF} 0.66}     & 2.12                        & 0.98                        \\
04                    & 6.64          & 4.02          & 2.96          & 2.58          & 3.76    & 1.07           & 1.59       & 0.71       & 0.71          & {\color[HTML]{3531FF} 0.50}          & 1.77           & /              & 2.84         & 3.07         & \textbf{0.51}   & 0.65     &  {\color[HTML]{3531FF} 0.65}                        & \textbf{0.45}               \\
05                    & 3.97          & 1.93          & 2.29          & 1.08          & {\color[HTML]{3531FF} 1.02}    & {\color[HTML]{3531FF} 0.54}            & 1.98       & 0.92       & \textbf{0.57} & \textbf{0.38} & 2.35           & /              & /            & /            &   1.04           & 0.69     & 1.07                        & 0.59 \\
06                    & 1.95          & 1.59          & 1.77          & 1.00          & 0.92    & 0.46           & 0.92       & 0.46       & {\color[HTML]{3531FF} 0.65}          & {\color[HTML]{3531FF} 0.39}          & 1.88           & /              & /            & /            & 0.71            & 0.50     & \textbf{0.62}               & \textbf{0.34}               \\
07                    & 5.17          & 3.35          & 1.55          & 1.42          &  {\color[HTML]{3531FF} 0.64}    & \textbf{0.45}  & 1.04       & 0.73       & \textbf{0.63} & {\color[HTML]{3531FF} 0.50}          & 1.77           & /              & /            & /            & 1.70            & 0.89     & 1.86                        & 1.64                        \\
08                    & 10.04         & 4.93          & 4.42          & 2.14          &  {\color[HTML]{3531FF} 1.58}    & {\color[HTML]{3531FF} 0.75}           & 2.14       & 1.05       & \textbf{1.12} & \textbf{0.44} & 2.89           & /              & /            & /            & 2.12            & 0.77     & 2.04 & 0.97                        \\
09                    & 6.93          & 2.89          & 3.95          & 1.71          & 1.97    & 0.77           & 1.95       & 0.92       & {\color[HTML]{3531FF} 0.77}          & {\color[HTML]{3531FF} 0.48}          & 4.94           & /              & /            & /            & 1.37            & 0.58     & \textbf{0.63}               & \textbf{0.35}               \\
10                    & 8.91          & 4.74          & 6.13          & 2.60          & 1.31    & 0.62           & 3.46       & 1.28       & \textbf{0.79} & {\color[HTML]{3531FF} 0.57}          & 3.27           & /              & 2.41         & 6.70         & 1.80            & 0.93     & {\color[HTML]{3166FF} 1.18} & \textbf{0.45}               \\ \hline
aveage                & 7.36          & 3.41          & 4.74          & 1.93          & 1.92    & 0.73           & 2.12       & 0.91       & \textbf{0.84} & \textbf{0.46} & 3.15           & /              & /            & /            & 1.33            & 0.69     & {\color[HTML]{3166FF} 1.27} & {\color[HTML]{3166FF} 0.66} \\ \hline
\end{tabular}}
\label{tab:results}
\end{table*}

Table 1 shows the evaluation results of the mentioned methods on the KITTI dataset. We use $t_{rel}$: the Average Transnational RMSE($\%$) and $r_{rel}$: Average Rotational RMSE($^\circ$/100m) to evaluate the results of different methods. There are few deep learning-based approaches for LiDAR odometry estimation that have comparable results. DeepPCO\cite{DeepPCO} only reports the results on its validation sequences. However they did not specify the unit of $t_{rel}$ and $r_{rel}$. Based on the trajectories they provide, we determine that their $t_{rel}$ is 2.63 and $r_{rel}$ is 3.05 for sequence 04, and  $t_{rel}$ is 2.47 and $r_{rel}$ is 6.59 for sequence 10 with the same unit as in Table ~\ref{tab:results}. CAE-LO\cite{CAE-LO} 
did not provide the results on KITTI Seq 00-10. LO-Net\cite{LO-Net} is one of the best deep learning methods for LiDAR based odometry estimation. From Table ~\ref{tab:results}, The Seq 07 and 08 are not used to train \lodo{}. The bold number indicates the best performance among all the methods, and the blue number indicates the runner-up. In some sequences, our results are even better than LOAM. However, LOAM still remains the best option which is the state-of-art Geometry based approach. Until now there is no deep learning method can beat the LOAM algorithm, but it is clear that deep learning methods become more and more accurate. 

\begin{figure}[h]
\setlength{\abovecaptionskip}{-0.1cm} 
\setlength{\belowcaptionskip}{-0.1cm} 
  \centering
  \includegraphics[width=0.4\textwidth]{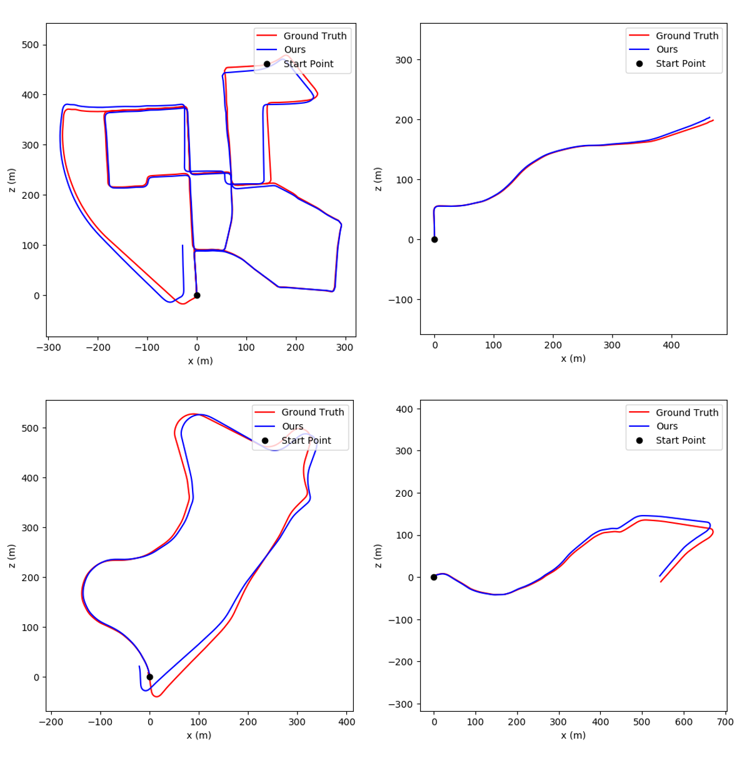}
  \caption{\footnotesize{2D estimate trajectory of our training sequences: KITTI Seq.00 (upper left), Seq.03 (upper right), Seq.09 (lower left), Seq.10 (lower right) with ground truth.}}
  \label{fig:train trajectory}
\end{figure}

\begin{figure}[h]
\setlength{\belowcaptionskip}{-0.1cm} 
     \centering
     \begin{subfigure}[b]{\columnwidth}
         \centering
         \includegraphics[width=0.82\textwidth]{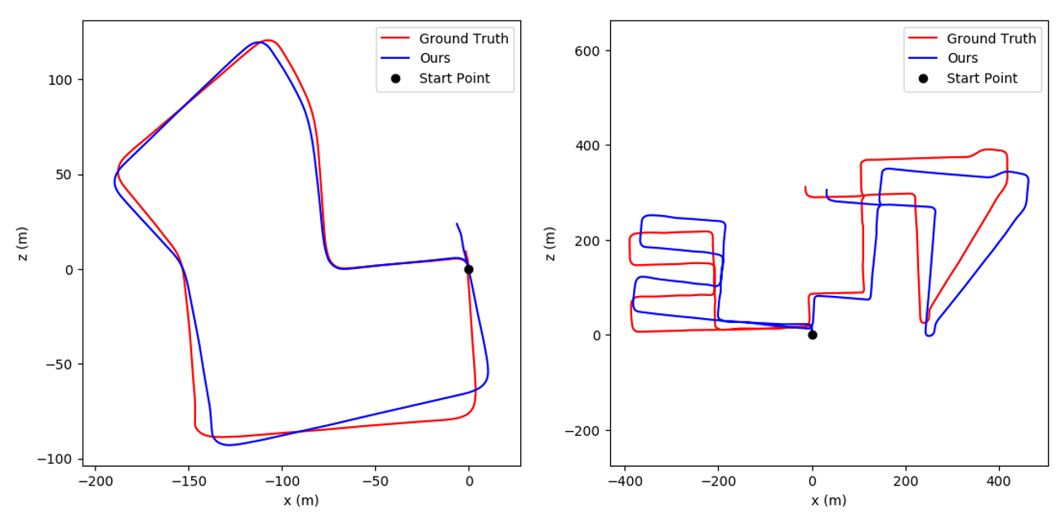}
         \captionsetup{font={scriptsize}} 
         \caption{Estimated trajectory plots of KITTI Seq. 07(Left) and 08(Right) with ground truth.}
         \label{fig:evaluation a}
     \end{subfigure}

     \begin{subfigure}[b]{\columnwidth}
         \centering
         \includegraphics[width=0.82\textwidth]{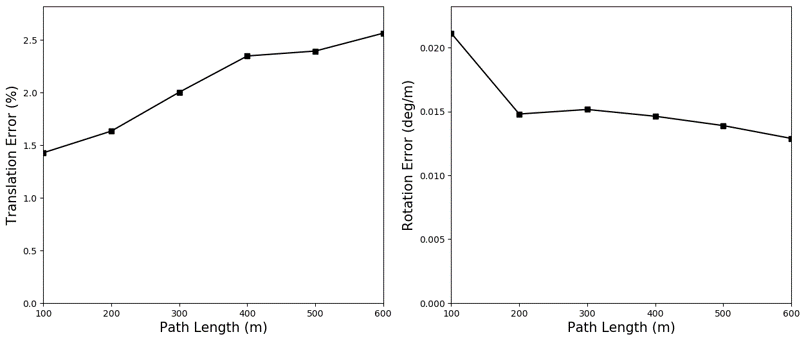}
         \captionsetup{font={scriptsize}} 
         \caption{The average errors of translation and rotation with respect to path length interval of KITTI Seq. 07}
         \label{fig:evaluation b}
     \end{subfigure}
     
     \begin{subfigure}[b]{\columnwidth}
         \centering
         \includegraphics[width=0.82\textwidth]{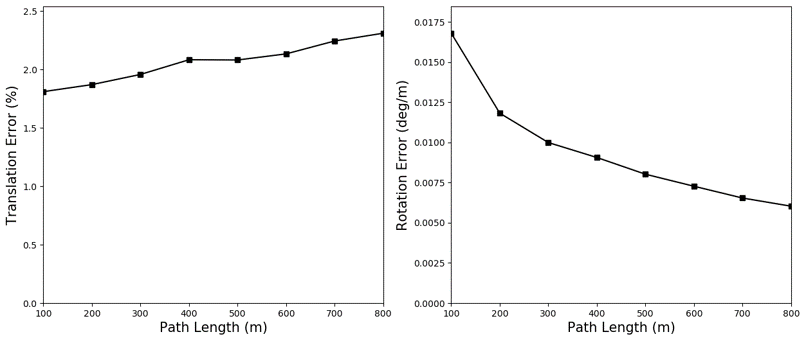}
         \captionsetup{font={scriptsize}} 
         \caption{The average errors of translation and rotation with respect to path length interval of KITTI Seq. 08}
         \label{fig:evaluation c}
     \end{subfigure}
     
        \caption{Evaluation of our estimation on our testing set:KITTI Seq. 07 and Seq. 08}
        \label{fig:evaluation}
\end{figure}

 Figure ~\ref{fig:train trajectory} shows our estimated 2D trajectory plots of our training sequences: KITTI sequence 00, 03, 09, and 10 with ground-truth. Figure ~\ref{fig:evaluation a} shows our estimated 2D trajectory plots of our testing sequences: KITTI sequence 07 and 08 with ground-truth. The blue line is our estimated trajectory and the red line is the ground truth trajectory.  Our \lodo{} can produce accurate pose estimation with respect to ground truth. The average errors of translation and rotation with respect to path length interval of KITTI sequence 07 and 08 are shown in the Figure ~\ref{fig:evaluation b} and  ~\ref{fig:evaluation c} respectively.

\subsection{Ablation study}
In this section, we investigate the effects of different factors of our odometry estimation on the KITTI dataset. We change the observation parameter while others remain as default parameters to evaluate our network.

\noindent
 \textbf{Rotation augmentation in Rotation Estimation Module}

In section \ref{section:train lodo}, we clam that our rotation data augmentation procedure contribute to estimate 3-DoF relative odometry rotation information. We list the results of KITTI Seq. 07 and 08 in Table \ref{tab:ablation study rotation} with different rotation augmentation ratio. It proves that our rotation data augmentation procedure can improve our network performance and when $r$ = 0.05 achieves the best performance. 

\begin{table}
  \caption{Comparison of different combinations of the rotation augmentation ratio.}
  \scalebox{0.8}{
\begin{tabular}{c|ccccc|cl}
\hline
                        & \multicolumn{5}{c|}{{ with augmentation}}                                                                                                                                               & \multicolumn{2}{c}{{ w/o}}          \\
                      & { ratio r}                     & { 0.03} & { 0.04} & { 0.05}          & { 0.06} & \multicolumn{2}{c}{{ augmentation}} \\ \hline
                                                      & \multicolumn{1}{c|}{{ t\_rel}} & { 2.99} & { 2.48} & { \textbf{1.86}} & { 2.09} & \multicolumn{2}{c}{{ 3.77}}         \\
\multirow{-2}{*}{Seq 07}                              & \multicolumn{1}{c|}{{ r\_rel}} & { 1.79} & { 1.58} & { \textbf{1.64}} & { 1.46} & \multicolumn{2}{c}{{ 2.18}}         \\ \hline
                        & \multicolumn{1}{c|}{{ t\_rel}} & { 3.46} & { 2.09} & { \textbf{2.04}} & { 4.59} & \multicolumn{2}{c}{{ 4.33}}         \\
\multirow{-2}{*}{{ Seq 08}} & \multicolumn{1}{c|}{{ r\_rel}} & { 1.47} & { 0.98} & { \textbf{0.97}} & { 2.01} & \multicolumn{2}{c}{{ 1.80}}          \\ \hline
\end{tabular}}
\label{tab:ablation study rotation}
\end{table}

\noindent
 \textbf{Number of MKPs selected by MKPs Selection Module}

As we state in section \ref{section:Implementation detail}, we extract 1000 MKPs on consecutive spherical images. We compare the results on KITTI Seq. 07 and 08 with different numbers as shown in the Table \ref{tab:k mkp}. When choosing 100 MKPs, the network achieves the best performance. 

\begin{table}
  \caption{Comparison of top $k$ MKPs choose by MKPs selection module.}
  \scalebox{0.8}{
\begin{tabular}{c|ccccc}
\multicolumn{1}{l|}{{ }}    & { Top k}                       & { 50}   & { 100}           & { 200}  & { 300}  \\ \hline
{ }                         & \multicolumn{1}{c|}{{ t\_rel}} & { 3.70} & { \textbf{1.86}} & { 2.97} & { 4.22} \\
\multirow{-2}{*}{{ Seq 07}} & \multicolumn{1}{c|}{{ r\_rel}} & { 2.84} & { \textbf{1.64}} & { 1.89} & { 2.28} \\ \hline
{ }                         & \multicolumn{1}{c|}{{ t\_rel}} & { 3.20} & { \textbf{2.04}} & { 5.38} & { 4.09} \\
\multirow{-2}{*}{{ Seq 08}} & \multicolumn{1}{c|}{{ r\_rel}} & { 1.65} & { \textbf{0.97}} & { 2.04} & { 2.00} \\ \hline
\end{tabular}}

\label{tab:k mkp}
\end{table}

\section{Conclusions}
In this paper,we present a novel deep learning-based LiDAR odometry estimation framework named LodoNet. Within the framework we propose a new approach that extract the matched keypoint pairs(MKPs) by applying conventional image-based feature describer from projected LiDAR images. With the help of PointNet, we adopt the MKPs to estimate the movements between the LiDAR frames, which finally result in the LiDAR odometry estimation. Experiments on KITTI dataset demonstrate the effectiveness of our framework compared with existing deep learning approaches. More over, since our framework is mainly integrated by a conventional feature describer and a light-weighed neural network, which can be easily deployed to the automated driving systems without allocating many computational resources. In our future work, we are going to explore how to further integrate the MKPs extraction and odometry estimation steps for a more accurate and efficient LiDAR odometry estimator.

\newpage

\bibliographystyle{ACM-Reference-Format}
\balance
\bibliography{sample-base}

\end{document}